\pdfoutput=1
\documentclass[11pt]{article}

\usepackage[]{emnlp2021}

\usepackage{times}
\usepackage{latexsym}

\usepackage[T1]{fontenc}
\usepackage[utf8]{inputenc}

\usepackage{microtype}

\usepackage{graphics}

\title{Finnish Dialect Identification: The Effect of Audio and Text }

\author{Mika Hämäläinen, Khalid Alnajjar, Niko Partanen and Jack Rueter \\
  Department of Digital Humanities\\
  University of Helsinki \& Rootroo Ltd\\
  \texttt{firstname.lastname@helsinki.fi} \\}

\begin{document}
\maketitle
\begin{abstract}
Finnish is a language with multiple dialects that not only differ from each other in terms of accent (pronunciation) but also in terms of morphological forms and lexical choice. We present the first approach to automatically detect the dialect of a speaker based on a dialect transcript and transcript with audio recording in a dataset consisting of 23 different dialects. Our results show that the best accuracy is received by combining both of the modalities, as text only reaches to an overall accuracy of 57\%, where as text and audio reach to 85\%. Our code, models and data have been released openly on Github and Zenodo.
\end{abstract}

\section{Introduction}

We present an approach for identifying the dialect of a speaker automatically solely based on text and on audio and text together. We compare the unimodal approach to the bimodal one. There are no previous dialect identification approaches for Finnish. There are several situations were a dialect identification method can be of use. For example, if we have ASR models fine tuned for specific dialects, the dialect identification from audio could be used as a preprocessing step. 
The model could also be used to label recorded materials automatically in order to create archival metadata. % Also there are certainly materials that are missing information about the dialect, or where the current dialect label is not correctly assigned. The latter happens easily for technical reasons already. Again, dialect identification method could be used to organize the content in the archives. 
In order to make our contribution useful for others, we have released our code, models and processed data openly on GitHub\footnote{\url{https://github.com/Rootroo-ltd/FinnishDialectIdentification}} and Zenodo\footnote{\url{https://zenodo.org/record/5330673}}.

Finnish is a large Uralic language that is one of the official languages of Finland, and is used essentially at all levels of the modern society. 
There are approximately five million Finnish speakers. 
The language belongs to the Finnic branch of the Uralic language family, and is very closely related to Karelian, Meänkieli and Kveeni, and is also closely related to the Estonian language. 
It is more distantly related to numerous Uralic languages spoken in Russia. 

The history of written Finnish starts in the 16th century. Current orthography is connected to this written tradition, which developed into the current form in the late 19th century with a conscious planning and systematic development of the lexicon. After this, the changes have been minor \cite[16]{hakkinen1994agricolasta}, and also impacted lexicon, especially what it comes to the development of the vocabulary of the modern society and traditional agrarian terminology becoming less known. 

The Finnish spoken language, however, is still largely based on Finnish dialects. 
In the 20th century some of the strongest dialectal features have been disappearing, but there are still clearly distinguishable spoken vernacular varieties that are regionally marked. 
It has been shown that instead of clear disappearance of dialects there are various features that are spreading, but not at uniform rate, and reportedly younger speakers use the areally marked features less than the older speakers \cite[92]{lappalainen2001sosiolingvistinen}. 
Finnish vernaculars also represent historically rather different Finnic varieties, with major split between Eastern and Western dialects. 
There are, however, also dialect continuums and traditionally found gradual differentiation from region to region. 

Many of the changes have been lexical due to technical innovations and modernization of the society: orthographic spelling conventions have largely remained the same. Spoken Finnish, on the other hand, traditionally represents an areally divided dialect continuum, with several sharp boundaries, and many regions of gradual differentiation from one municipality to another municipality. 

As mentioned, in the later parts of the 20th century relatively strong dialect leveling has been taking place. Some of the Finnish dialects may already be concerned endangered, although the complex relationship between contemporary vernaculars and the most traditional dialectal forms makes this hard to ascertain. Dialect leveling in itself is a process known from many parts of Europe \cite{auer2018dialect}.  However, in the case of Finnish the written standard has remained relatively far from the spoken Finnish, besides individual narrow domains such as news broadcasts were the written form is used also in speech. 

Additionally there have been distinct text collections that include materials from this dialect archive. These include dialect books specific regions and municipalities, such as Oulun murrekirja [Dialect Book of Oulu] \cite{paakkonen1994oulun} or Savonlinnan seudun murrekirja [Dialect book of Savonlinna region] \cite{palander1986a}. There have also been more recent larger collections that contains excerpts from essentially all dialects \cite{lyytikainen2013suomen}. %Especially with the current OCR technologies and forced alignment tools combining these resources into digitally available corpora is also becoming possible, but is not something that would be currently undertaken. However, this all contributes to the body of work that exists for the Finnish dialects, especially in the form they were spoken in early 20th century. 

Especially in the later parts of 21th century the spoken varieties have been leveling away from very specific local dialects, and although regional varieties still exist, most of the local varieties have certainly became endangered. Similar processes of dialect convergence have been reported from different regions in Europe, although with substantial variation \cite{auer2018dialect}. In the case of Finnish this has not, however, resulted in merging of the written and spoken standards, but the spoken Finnish has remained, to our day, very distinct from the written standard. In a late 1950s, a program was set up to document extant spoken dialects, with the goal of recording 30 hours of speech from each municipality. This work resulted in very large collections of dialectal recordings \cite[448-449]{lyytikainen1984suomen}. Many of these have been published, and some portion has also been manually normalized. Dataset used is described in more detail in Section 3 Data.

In Finnish linguistics the dialect identification has primarily been studied in the context of folk linguistics. 
In this line of research the perceptions of native speakers are investigated \cite{niedzielski2000folk}. 
This type of studies have been done for Finnish, for example, by \citet{mielikainenEtAl2014a}, \citet{rasanen2015kansandialektologinen} and \citet{palander2011a}. 
It has been possible to suggest for individual dialects which features are the most stable and will remain as local regional markers, and which seem to be in retention \cite[25]{rasanen2015kansandialektologinen}. 
In this study we conduct just individual experiments and report their results, but in the further research we hope the results could be analyzed in more detail in connection with the earlier dialect perception studies, as we believe the differences in perceived dialect differences could be compared to the difficulties and successes the model has to differentiate individual varieties. 

% Niko: I think this was more relevant in the normalization paper, to understand why the normalization could work, but for dialect identification it has to be reworded. And it shouldn't be 100% same as in the generation paper.
%Finnish orthography is largely phonemic within the language variety used in that representation, although, as discussed above, the relationship to actual spoken Finnish is complicated. Phonemicity of the orthography is still a very important factor here, as the differences between different varieties are mainly displaying historically developed differences, and not orthographic particularities that would be essentially random from contemporary point of view. Thereby the differences between Finnish dialects, spoken Finnish and Standard Finnish are highly systematic and based to historical sound correspondences and sound changes, instead of more random adaptation of historical spelling conventions that would be typical for many languages. Due to the phonemicity of the Finnish writing system, dialectal differences are also reflected in informal writing. People speaking a dialect oftentimes also write it as they would speak it when communicating with friends and family members. %This is different from English in that, for example, although Australians and Americans pronounce the word \textit{today} differently, they would still write the word in the same way. In Finnish, such a dialectal difference would result in a different written form as well.

\begin{table}[!h]
\centering
\footnotesize
\begin{tabular}{|l|l|l|}
\hline
Dialect & Short & Sentences \\ \hline
Etelä-Häme & EH & 1860 \\ \hline
Etelä-Karjala & EK & 813 \\ \hline
Etelä-Pohjanmaa & EP & 2684 \\ \hline
Etelä-Satakunta & ES & 848 \\ \hline
Etelä-Savo & ESa & 1744 \\ \hline
Eteläinen Keski-Suomi & EKS & 2168 \\ \hline
Inkerinsuomalaismurteet & IS & 4035 \\ \hline
Kaakkois-Häme & KH & 8026 \\ \hline
Kainuu & K & 3995 \\ \hline
Keski-Karjala & KK & 1640 \\ \hline
Keski-Pohjanmaa & KP & 900 \\ \hline
Länsi-Satakunta & LS & 1288 \\ \hline
Länsi-Uusimaa & LU & 1171 \\ \hline
Länsipohja & LP & 1026 \\ \hline
Läntinen Keski-Suomi & LKS & 857 \\ \hline
Peräpohjola & P & 1913 \\ \hline
Pohjoinen Keski-Suomi & PKS & 733 \\ \hline
Pohjoinen Varsinais-Suomi & PVS & 3885 \\ \hline
Pohjois-Häme & PH & 859 \\ \hline
Pohjois-Karjala & PK & 4292 \\ \hline
Pohjois-Pohjanmaa & PP & 1801 \\ \hline
Pohjois-Satakunta & PS & 2371 \\ \hline
Pohjois-Savo & PSa & 2344 \\ \hline
\end{tabular}
\caption{Dialects and the number of sentences in each dialect in the corpus}
\label{tab:dialects-sentences}
\end{table}

\section{Related work}

The current approaches to Finnish dialect have focused on the textual modality only. Previously, bi-directional LSTM (long short-term memory) based models have been used to normalize Finnish dialects to standard Finnish \cite{partanen2019dialect} and to adapt standard Finnish text into different dialectal forms \cite{hamalainen2020automatic}. Similar approach has also been used to normalize historical Finnish \cite{hamalainen-etal-2021-lemmatization,partanen-etal-2021-linguistic}.

The closest research to our paper conducted for Finnish has been detection of foreign accents from audio. \citet{behravan2013foreign} have detected foreign accents from audio only by using i-vectors. However, foreign accent detection is a very different task to native speaker dialect detection. Many foreign accents have clear cues through phonemes that are not part of the Finnish phonotactic system, where as with dialects, all phonemes are part of Finnish.

There have been several recent approaches for Arabic to detect dialect from text \cite{appiah-balaji-b-2020-semi,talafha2020multi,alrifai2021arabic}. Textual dialect detection has been done also for German \cite{jauhiainen2018heli}, Romanian \cite{zaharia2021dialect} and Low Saxon \cite{siewert-etal-2020-lsdc}. The methods used range from traditional machine learning with features such as n-grams to neural models with pretrained embeddings, as it is typically the case in NLP research. None of these approaches use audio, as they rely on text only.

At the same time, North Sami dialects have been identified from audio by training several models, kNNs, SVMs, RFs, CRFs, and LSTM, based on extracted features \cite{kakouros2020dialect}. \citet{9180347} use Mel-weighted SFF spectrogram to detect spoken Arabic dialects. Mel spectograms are also used by \citet{10.1145/3411109.3411123}. All these approaches are mono-modal and use only audio. % None of these approaches consider text, as they only work on audio.

Based on our literature review, the existing approaches use either text or audio for dialect detection. We, however, use both modalities and apply them on a language with no existing dialect detection models.

\section{Data}

The Finnish dialects are exceptionally well documented. 
In the 1950s the Finnish dialect archive was formed with the goal of recording 30 hours of speech from each Finnish municipality. 
This goal was reached fast, and exceeded, resulting in a very large collection of archived materials that is stored in the Institute for the Languages of Finland \cite[448-449]{lyytikainen1984suomen}, and known as \textit{Tape Archive of the Finnish Language}\footnote{\url{https://www.kotus.fi/en/corpora_and_other_material/spoken_language_corpora}}. 
There have been numerous publications based on these materials, although it is hard to estimate into which extent this covers the entire body of recorded work, which totals 24,000 hours of audio.

The largest individual publication of these materials is beyond doubt the \textit{Samples of Spoken Finnish} series that was published in 1978–2000 as 50 booklets.\footnote{\url{https://www.kotus.fi/aineistot/puhutun_kielen_aineistot/murreaanitteita/suomen_kielen_naytteita_-sarja}} 
Each book contained approximately two hours of transcriptions, from two different speakers, and represents a different municipality. 
Later these materials have been digitized and published as an openly licensed dialect corpus \cite{skn}. 
%Another openly available corpus is The Finnish Dialect Syntax Archive Corpus \cite{LA}, which partially contains also recordings archived in the Tape Archive of the Finnish Language, but also materials from the University of Turku. 
There are also other related corpora, most importantly \textit{The Finnish Dialect Syntax Archive} that contains similar recordings annotated morphosyntactically \cite{LA}. 
Since 1980s follow-up research has been done in selected municipalities to track the changes in the dialects \cite[413]{lyytikainen2010suomen}, which is another significant line of research that complements these older dialect materials.  

%Finnish dialect materials have been collected systematically since late 1950s. These materials are currently stored in the Finnish Dialect Archive within Institute for the Languages of Finland, and they amount all in all 24,000 hours. The initial goal was to record 30 hours of speech from each pre-war Finnish municipality. This goal was reached in the 70s, and the work evolved toward making parts of the materials available as published text collections. Another approach that was initiated in the 80s was to start follow-up recordings in the same municipalities that were the targets of earlier recording activity. 

Later the work on these published materials has resulted in multiple electronic corpora that are currently available. Although they represent only a tiny fraction of the entire recorded material, they reach remarkable coverage of different dialects and varieties of spoken Finnish. Some of these corpora contain various levels of manual annotation, while others are mainly plain text with associated metadata. Materials of this type can be characterized by an explicit attempt to represent dialects in linguistically accurate manner, having been created primarily by linguists with formal training in the field. These transcriptions are usually written with a transcription systems specific for each research tradition. The result of this type of work is not simply a text containing some dialectal features, but a systematic and scientific transcription of the dialectal speech.

The corpus we have used in this study is the above-mentioned Samples of Spoken Finnish corpus \cite{skn}. The electronic version contains manually annotated normalization into standard Finnish. The corpus is almost 700,000 tokens large. 
The digital version, including audio, is published with CC-BY license, and is available in the Language Bank of Finland.\footnote{\url{http://urn.fi/urn:nbn:fi:lb-201407141}} 
We have selected it into this study because of the open license and large dialectal scope. 
We have downloaded the corpus with the original audio files, and extracted from the audio all utterances that are shorter than 10 seconds in length. 
The dialect region classification is taken directly from the corpus metadata. 

Despite the successful attempt of the authors of the corpus to include all dialects, the dialects are not entirely equally represented in the corpus. One reason for this is certainly the different sizes of the dialect areas, and the variation introduced by different speech rates of individual speakers. The difference in the number of sentences per dialect can be seen in Table \ref{tab:dialects-sentences}. We do not consider this uneven distribution to be a problem, as it is mainly a feature of this dataset. The data has been tokenized and the dialectal transcriptions are aligned with audio on a sentence level. This makes our task with the dialect detection model easier as no alignment is required. We randomly sort the sentences in the data and split them into a training (70\% of the sentences), validation (15\% of the sentences) and test (15\% of the sentences) sets. This means that the models are trained and tested on a sentence level rather than on smaller chunks.

\section{Dialect detection}

In this section, we describe the two different models we used to detect dialect automatically in the corpus. The first method is based on text only and the second method uses text and audio. Both of the methods used the same training, validation and test splits.

\subsection{Text only model}

We train a dialect classification model using a bi-directional long short-term memory (LSTM) based model \cite{hochreiter1997long} by using OpenNMT-py \cite{opennmt} with the default settings except for the encoder where we use a BRNN (bi-directional recurrent neural network) \cite{schuster1997bidirectional} instead of the default RNN (recurrent neural network), since BRNN based models have been shown to provide better results in a variety of tasks. 

We use the default of two layers for both the encoder and the decoder and the default attention model, which is the general global attention presented by  \citet{luong2015effective}. The models are trained for the default of 100,000 steps. The model receives dialectal text\footnote{We also experimented with a character-level model using the same neural network structure, but the accuracy remained low, only 37\%} as input and predicts a dialect name as an output.

\subsection{Text and audio model}

Our multi-modal model makes use of the dialectal text and audio. The model combines BERT \cite{devlin2019bert} and XLSR-Wav2Vec2 \cite{baevski2020wav2vec} neural models trained on Finnish data. We utilize the uncased Finnish BERT model\footnote{\url{https://huggingface.co/TurkuNLP/bert-base-finnish-uncased-v1}} \cite{virtanen2019multilingual}. The multilingual XLSR-Wav2Vec2 model released by Facebook does not support Finnish. Therefore we use a Finnish XLSR-Wav2Vec2 model\footnote{\url{https://huggingface.co/aapot/wav2vec2-large-xlsr-53-finnish}} that is fine-tuned using readily available Finnish audio datasets: Finnish Common Voice \cite{ardila-etal-2020-common}, CSS10 Finnish \cite{park2019css10} and Finnish parliament session 2\footnote{\url{ http://urn.fi/urn:nbn:fi:lb-2017020201}} for 30 epochs. All audio input is resampled to 16kHz.% https://huggingface.co/aapot/wav2vec2-large-xlsr-53-finnish

Our multi-modal model follows a siamese neural network architecture, where one side of the network is dedicated to text and the other to audio. We ensure that both sides produce an equal size of features by 1) setting a fixed input length to BERT where padding and truncating is applied where necessary and 2) having two average pooling layers following the output of each side. For the textual output, a global average pooling is applied, whereas an adaptive average pooling is applied to the audio output. Afterwards, the pooled output is concatenated and followed by a dropout layer with a probability of 20\%. Lastly, a fully connected dense layer is employed as the classification layer. In total, the network has 439 million trainable parameters and we fine-tuned it for 3 full epochs with a learning rate of 1e-4.

\section{Results}

The results of the two models can be seen in Table \ref{tab:resluts}. These results were calculated using scikit-learn\footnote{sklearn.metrics.classification\_report} \cite{scikit-learn}. It is clear from the results that the text only model performed worse for every single dialect than the audio and text model. In terms of overall accuracies, the text based model reached only an accuracy of 57\%, where as the text and audio based model reach to an accuracy of 85\%. 
This indicates that the audio has classificatory features that are not represented in the text version alone, although the text is in a transcription system that accurately captures various dialectal phenomena. 

\begin{table}[ht]
\small
\centering
\resizebox{0.47\textwidth}{!}{%
\begin{tabular}{|l|l|l|l|l|l|l|}
\hline
    & \multicolumn{3}{c|}{bi-LSTM on text} & \multicolumn{3}{c|}{Audio + BERT} \\ \hline
    & precison   & recall  & f1    & precision    & recall    & f1     \\ \hline
EH  & 0.49       & 0.48    & 0.49  & 0.97         & 0.89      & 0.93   \\ \hline
EK  & 0.51       & 0.44    & 0.47  & 0.86         & 0.57      & 0.69   \\ \hline
EP  & 0.72       & 0.67    & 0.69  & 0.68         & 0.93      & 0.79   \\ \hline
ES  & 0.5        & 0.53    & 0.51  & 0.79         & 0.82      & 0.8    \\ \hline
Esa & 0.38       & 0.37    & 0.38  & 0.6          & 0.97      & 0.74   \\ \hline
EKS & 0.44       & 0.38    & 0.41  & 0.9          & 0.85      & 0.87   \\ \hline
IS  & 0.74       & 0.75    & 0.75  & 0.96         & 0.86      & 0.91   \\ \hline
KH  & 0.67       & 0.74    & 0.7   & 0.86         & 0.97      & 0.92   \\ \hline
K   & 0.53       & 0.49    & 0.51  & 0.97         & 0.83      & 0.9    \\ \hline
KK  & 0.57       & 0.54    & 0.56  & 0.92         & 0.95      & 0.93   \\ \hline
KP  & 0.46       & 0.45    & 0.46  & 0.81         & 0.87      & 0.84   \\ \hline
LS  & 0.47       & 0.38    & 0.42  & 0.98         & 0.74      & 0.84   \\ \hline
LU  & 0.56       & 0.52    & 0.54  & 0.97         & 0.98      & 0.98   \\ \hline
LP  & 0.34       & 0.32    & 0.33  & 0.94         & 0.92      & 0.93   \\ \hline
LKS & 0.34       & 0.46    & 0.39  & 0.72         & 0.99      & 0.83   \\ \hline
P   & 0.55       & 0.58    & 0.56  & 0.71         & 0.93      & 0.81   \\ \hline
PKS & 0.4        & 0.38    & 0.39  & 0.93         & 0.62      & 0.75   \\ \hline
PVS & 0.75       & 0.72    & 0.73  & 0.91         & 0.74      & 0.82   \\ \hline
PH  & 0.32       & 0.31    & 0.31  & 0.83         & 0.63      & 0.72   \\ \hline
PKS & 0.6        & 0.58    & 0.59  & 0.92         & 0.8       & 0.86   \\ \hline
PP  & 0.4        & 0.45    & 0.42  & 0.74         & 0.38      & 0.5    \\ \hline
PS  & 0.5        & 0.53    & 0.51  & 0.9          & 0.89      & 0.89   \\ \hline
PSa & 0.43       & 0.47    & 0.45  & 0.68         & 0.87      & 0.76   \\ \hline
\end{tabular}
}
\caption{Results for the two models}
\label{tab:resluts}
\end{table}

When comparing the per dialect performance of the better model with the amount of data available for each dialect, we can make an interesting observation that a high amount of data does not equal to a high F1-score. Out of the 10 dialects with the largest amount of samples in the data, only 3, \textit{Kaakkois-Häme}, \textit{Inkerinsuomalaismurteet} and \textit{Kainuu}, reached to an F1-score of at least 0.90. The F1-score of the dialect with the second highest number of samples, \textit{Pohjoinen Keski-Suomi}, was only 0.86. Other dialects that had an F1-score of at least 0.9 were the 11th most resourced \textit{Etelä-Häme}, the 14th most resourced \textit{Keski-Karjala} and the 16th and 17th most resourced \textit{Länsi-Uusimaa} and \textit{Länsipohja}.

The lowest F1-score was 0.5 for \textit{Pohjois-Pohjanmaa}. This is interesting as the dialect is the 12th most resourced one. Even the two least resourced dialects in our dataset, \textit{Etelä-Karjala} and \textit{Pohjoinen Keski-Suomi} got higher F1-scores, 0.69 and 0.75 respectively. These results are an indication that some of the dialects are more clearly marked making them easier to detect even with less data, while some other dialects may have undergone a process of dialect leveling (see \citealt{hinskens1998dialect}) making them less distinct from other dialectal forms of Finnish. It is also possible that some dialects are already significantly close to one another, and thereby the model simply cannot distinguish them accurately. Further error analysis could reveal important details of this type. 

\section{Conclusions}

We have presented the first model for Finnish dialect classification for a relatively large number of different dialects, 23 in total. Based on our experiments, a text only model is not as effective in dialect classification as a model with text and audio. It is clear that the amount of data alone is not the only variable that constitutes a high performance of the model for a given dialect, but also how distinctive a given dialect is from other dialects. 
Since the speakers in the test set were not present in the training, we are confident that the dialect is the feature that the model has learned to predict. 

Using the audio materials offers in itself new interesting possibilities for dialect clustering and comparison. 
Traditional dialect atlases have also been used in automatic comparison and grouping of different Finnish dialects \cite{syrjanen2016applying}. 
In further research we believe also this kind of information could be connected to the analysis to show how the dialect identification exactly interacts with the dialectal variation and differences at close municipality level. 
At the same time the identifiability of a dialect must be connected to the degree of dialect leveling, linguistic distances and differences between them, so applying the model into newer recordings could also yield information about these processes.  

We have made all the data, code and models openly available on Github\footnote{\url{https://github.com/Rootroo-ltd/FinnishDialectIdentification}} and Zenodo\footnote{\url{https://zenodo.org/record/5330673}}. We believe that this is the only way to ensure this line of research continues for the Finnish language in the future as well.

\bibliography{anthology,custom}
\bibliographystyle{acl_natbib}

%jack: you've seen how our Murre goes right through Finnish dialect texts generating and analyzing, now meet Snarl (Haukku?) which takes us to a speech recognition level for analysis of audio clips in various Finnish vernaculars.

\end{document}